\documentclass[runningheads]{llncs}
\usepackage[T1]{fontenc}
\usepackage{graphicx}
\usepackage[hidelinks]{hyperref}
\usepackage{color}

\usepackage{amsmath,amssymb,mathtools}
\usepackage{algorithm}
\usepackage{tabularx}

\usepackage[capitalize]{cleveref}
\newcommand\tab[1][2em]{\hspace*{#1}}

\newcommand{\Pop}{\mathcal{N}}
\newcommand{\Cases}{\mathcal{C}}

\newcommand{\R}{\ensuremath{\mathbb{R}}}
\newcommand{\N}{\ensuremath{\mathbb{N}}}

\newcommand{\eps}{\varepsilon}
\let\doendproof\endproof
\renewcommand\endproof{~\hfill\qed\doendproof}

\begin{document}
\title{Population Diversity Leads to Short Running Times of Lexicase Selection}
%
%
\author{Thomas Helmuth\inst{1} \and
Johannes Lengler\inst{2} \and
William La~Cava\inst{3}$^*$} 
\authorrunning{T. Helmuth et al.}
%
\institute{Hamilton College, Clinton, New York, USA\\
\email{thelmuth@hamilton.edu} \and
ETH Z\"urich, Z\"urich, Switzerland \and
Boston Children's Hospital, Harvard Medical School, Boston, MA, USA\\
$^*$corresponding author: \email{william.lacava@childrens.harvard.edu}\\
}
\maketitle              
%
\begin{abstract}
In this paper we investigate why the running time of lexicase parent selection is empirically much lower than its worst-case bound of $O(N \cdot C)$. We define a measure of population diversity and prove that high diversity leads to low running times $O(N + C)$ of lexicase selection.
We then show empirically that genetic programming populations evolved under lexicase selection are diverse for several program synthesis problems, and explore the resulting differences in running time bounds.

\end{abstract}

\section{Introduction}

Semantic selection methods have been of increased interest as of late in the evolutionary computation community~\cite{liskowskiComparisonSemanticawareSelection2015,vanneschiSurveySemanticMethods2014} due to the observed improvements over more traditional selection methods (such as tournament selection) that only consider individual behavior in aggregate. 
One such method is lexicase selection~\cite{spectorAssessmentProblemModality2012,Helmuth:2015:ieeeTEC}, a parent selection method originally proposed for genetic programming.
Since then, the original algorithm and its variants have found success in different domains, including program synthesis~\cite{Helmuth:2015:GECCO}, symbolic regression~\cite{orzechowskiWhereAreWe2018b,lacavaContemporarySymbolicRegression2021a}, evolutionary robotics~\cite{moore:2018:alife}, and learning classifier systems~\cite{aenuguLexicaseSelectionLearning2019}.  

Although an active research community has illuminated many aspects of lexicase selection's behavior via experimental analyses~\cite{helmuthImpactHyperselectionLexicase2016,helmuthEffectsLexicaseTournament2016,moore:2018:alife,Helmuth:2020:GPEM}, theoretical analyses of lexicase selection have been slower to develop. 
Previous theoretical work has looked at the probability of selection under lexicase, and also made connections between lexicase selection and Pareto optimization~\cite{lacavaProbabilisticMultiobjectiveAnalysis2019}. 
A study focusing on ecological theory provided insights into the efficacy of lexicase selection~\cite{Dolson:2018:GECCOcomp}.
Additionally, the running time of a simple hill climbing algorithm utilizing lexicase selection has been analyzed for the bi-objective leading ones trailing zeroes benchmark problem~\cite{Jansen:2018:PPSN}.
However, the recursive nature of lexicase selection, and its step-wise dependence on the behavior of subsets of the population, make it difficult to analyze.  

We focus this paper on a particular gap in the theory of lexicase selection, which is an understanding of its running time. 
Although the worst-case complexity is known to be $O(N\cdot C)$, where $N$ is the population size and $C$ is the set of training cases, empirical data suggest the worst-case condition is extremely rare in practice~\cite{lacavaProbabilisticMultiobjectiveAnalysis2019}. 
Our goal is to explain this discrepancy through a combination of theory and experiment.

\subsection{Our contributions}
We find that the observed running time of lexicase selection can be explained with \emph{population diversity}, by which we mean the phenotypic/behavioral diversity of individuals in a population. Our contributions are threefold:
\begin{enumerate}
    \item We introduce a new way of measuring population diversity, ~\cref{def:diversity,def:diversity2}, which we call \emph{$\eps$-Cluster Similarity}, or \emph{$\eps$-Similarity} for short. Here, for different values of the parameter $\eps$, we obtain a measure of how similar the population is, where small $\eps$-Cluster Similarity corresponds to high diversity. As we show, this measure is not directly tied to other measures of diversity like the average phenotypical distance (\cref{sec:eps_cluster}) or the mean of the behavioral covariance matrix (\cref{fig:cov}). 
    \item We prove mathematically that lexicase selection is fast when applied to populations which are diverse. More precisely, we show that with low $\eps$-Cluster Similarity the expected running time of lexicase selection drops from $O(N\cdot C)$ to $O(N + C)$, where the hidden constants depend on the parameter $\eps$ and on the quantity $k$ that measures $\eps$-Cluster Similarity.
    \item Finally, we show empirically for several program synthesis problems~\cite{Helmuth:2015:GECCO} that genetic programming populations are indeed diverse in our sense (have low $\eps$-Similarity). We investigate which parameter $\eps$ gives the best running time guarantees for lexicase selection, and we find that the running time guarantees are substantially better than the trivial running time bound of $N\cdot C$. 
\end{enumerate}

Our findings apply to both discrete and continuous problems and population behaviors.  
Although we restrict our analysis to vanilla lexicase selection, we note the results generalize to other variants, including $\epsilon$-lexicase selection~\cite{lacavaEpsilonLexicaseSelectionRegression2016}\footnote{Our results hold for the original variant, later dubbed ``static" $\eps$-lexicase selection~\cite{lacavaProbabilisticMultiobjectiveAnalysis2019}.} and down-sampled lexicase selection~\cite{Helmuth:2020:ALife:downsampledlexicase}.
%




\section{Preliminaries}

\subsection{Lexicase Selection}

Lexicase selection is used to select a parent for reproduction in a given population. Unlike many common parent selection methods, lexicase selection does not aggregate an individual's performance into a single fitness value. Instead, it considers the loss (errors) on different training \textit{cases} (a.k.a. samples/examples) independently, never comparing (even indirectly) the results on one training case with those on another.


Lexicase selection begins by putting the entire population into a candidate pool. As a preprocessing step, all phenotypical \emph{duplicates} are removed from the pool, i.e., if several individuals give the same loss on all training cases, all but one are removed in the following filtering steps. Then lexicase selection repeatedly selects a new training case $t$ at random, and removes all individuals from the current candidate pool that do not achieve the best loss on case $t$ within the current pool. This process is repeated until the candidate pool contains only a single individual. 
If the remaining individual has phenotypical duplicates, the selected parent is taken at random from among these behavioral clones. 
We formalize the algorithm in Alg.~\ref{alg:lex}.

\begin{algorithm}[t]
\caption{{\bf Lexicase Selection} applied to a population $\Pop$ without duplicates, with discrete loss/error $L(n,c)$ on training cases $c \in \Cases$ and individual $n\in \Pop$. Returns an individual selected to be a parent.
$\Pop_t$ is the remaining candidate pool at step $t$, $\Cases'$ is the set of remaining training cases. 
}\label{alg:lex}
\noindent{
\small
\begin{tabularx}{\textwidth}{p{20em}}
\bf LEX($\Pop$, $\Cases$, $L$):         
\\
\tab    $\Cases' \leftarrow \Cases;\ t \leftarrow 0;\ \Pop_0 \leftarrow \Pop;$                                           
\\
\tab    \textbf{while} $|\Pop_t|>1$:                  
\\
\tab    \tab    $c \leftarrow$ random choice from $\Cases'$            \\
\tab    \tab    $\ell^* \leftarrow \min\{L(n,c) \mid n\in \Pop_t\}$    \\
\tab    \tab    $\Pop_{t+1} \leftarrow \{n\in \Pop_t \mid L(n,c) = \ell^*\}$                                    
\\
\tab    \tab    $\Cases' \leftarrow \Cases' \setminus \{c\}$           \\
\tab   \tab     $t \leftarrow t +1$                                    \\
\tab    \textbf{return} unique element from $\Pop_t$                \end{tabularx}
}
\end{algorithm}

We remark that the process is guaranteed to end up with a candidate pool of size one: whenever the candidate pool contains at least two individuals, they are not duplicates due to preprocessing. Hence, they differ on at least one training case $c$, and one of them is filtered out when $c$ is considered. So after all training cases have been processed, it is not possible that the candidate pool contains more than one individual.

Note that the described procedure selects a single individual from the population. In order to gather enough parents for the next generation, it is typically performed $O(N)$ times, where $N$ is the population size. 
An exception is the preprocessing step that only needs to be performed once each generation. 
Moreover, finding duplicates can be efficiently implemented via a hash map. 
Thus preprocessing is usually not the bottleneck of the procedure, and we will focus in this paper on the remaining part: the repeated reduction of the selection pool via random training cases. 
To exclude the effect of preprocessing, we will assume that the initial population is already free of duplicates.

In case of real-valued (non-discrete) losses, one typically uses a variant known as $\epsilon$-lexicase selection~\cite{lacavaEpsilonLexicaseSelectionRegression2016}. 
(Note this use of $\epsilon$ is distinct from that used in \cref{sec:eps_cluster}).  
In the original algorithm, later dubbed ``static" $\epsilon$-lexicase selection~\cite{lacavaProbabilisticMultiobjectiveAnalysis2019}, phenotypic behaviors are binarized prior to lexicase selection, such that individuals within $\epsilon$ of the population-wide minimal loss on $c$ have an error of 0, and otherwise an error of 1. 
Our results extend naturally to this version of $\epsilon$-lexicase selection. 

In contrast, in the ``dynamic" and ``semi-dynamic" variants of $\epsilon$-lexicase selection, lexicase selection removes those individuals whose loss is larger than $\ell^*+\epsilon$, where $\ell^*$ is the minimal loss in the \textit{current candidate pool}~\cite{lacavaProbabilisticMultiobjectiveAnalysis2019}. 
Our results may extend to these scenarios, but the framework becomes more complicated. 
Here, it is no longer possible to separate the preprocessing step (i.e., de-duplication) from the actual selection mechanism. 
Of course, it is still possible to define two individuals $n_1,n_2\in \Pop$ as \emph{duplicates} if they differ by at most $\delta$ on all training cases. 
But this is no longer a transitive relation, i.e., it may happen that $n_1$ and $n_2$ are duplicates, $n_2$ and $n_3$ are duplicates, but $n_1$ and $n_3$ are not duplicates. 
For these reasons, it is necessary to handle duplicates indirectly during the execution of the algorithm. 
To avoid these complications, we only present Alg.~\ref{alg:lex} in the case of discrete losses and without duplicates, but we do include the case of real-valued losses in our analysis.


The worst-case running time of lexicase selection is $O(N\cdot C)$, where $N:= |\Pop|$ is the population size and $C := |\Cases|$ is the number of training cases. The problem is that in an iteration of the while-loop, it may happen that $\Pop_{t+1} = \Pop_t$, i.e., that the candidate pool does not shrink. This is more likely for binary losses. Then, it may happen that $\Pop_{t} = \Pop$ for many iterations of the while-loop, and then computing $\ell^*$ and $\Pop_{t+1}$ needs time $O(N)$. Since the while-loop is executed up to $C$ times, this leads to the worst-case runtime $O(N\cdot C)$. Recall that we usually want to run the procedure $O(N)$ times to select all parents for the next generation, which then takes time $O(N^2\cdot C)$.

One might hope that the expected runtime is much better than the worst-case runtime. This is the case for many classical algorithms like quicksort, but for populations with an unfavorable loss profile, it is not the case here. Consider a population of individuals which have the same losses on all training cases except a single case $c$, and on case $c$ they all have different losses. Then the candidate pool does not shrink before this case $c$ is found, and finding this case needs $C/2$ iterations in expectation. Thus the expected runtime is still of order $O(N\cdot C)$.\footnote{This worst-case example does not hold if the losses are binary, but even that does not help much. It is possible to construct a population of $N$ individuals without duplicates that differ only on $\log_2 N$ binary training cases, and are identical on all other training cases. In this situation, the candidate pool does not shrink before at least one of those training cases is found, and in expectation this takes $C/\log_2 N$ iterations. Thus the expected runtime in this situation is at least $O(N\cdot C/\log N)$, which is not much better than $O(N\cdot C)$.}

So in order to give better bounds on the running time of lexicase selection, it is required to have some understanding of the involved populations. This is precisely the contribution of this paper: we define a notion of diversity that provably leads to a small running time of lexicase selection, and we empirically show that populations in genetic programming are diverse with respect to this measure.

\subsection{$\eps$-Cluster Similarity}\label{sec:eps_cluster}

We now come to our first main contribution, a new way of measuring diversity. The measure is in phenotype space, so it measures for a training set $\Cases$ how similar the individuals perform on this training set. We first introduce a useful notion, which is the \emph{phenotypical distance} of two individuals.


\begin{definition}[Phenotypical Distance]
Consider two individuals $m,n$ that are compared on a set $\Cases$ of training cases. The \emph{phenotypical distance} between $m$ and $n$ is the number of training cases in which $m$ and $n$ have different losses.

If the losses are real-valued, then for $\delta >0$, the \emph{phenotypical $\delta$-distance} between $m$ and $n$ is the number of training cases in which the losses of $m$ and $n$ differ by at least $\delta$.
\end{definition}

\begin{definition}[$\eps$-Cluster Similarity]\label{def:diversity}
Let $\Pop$ be a population of individuals, and $\Cases$ be a set of training cases with discrete losses, for example binary losses. Let $\eps \in [0,1]$. Then the \emph{$\eps$-Cluster Similarity} is defined to be the minimal $k\ge 2$ such that among every set of $k$ different individuals in $\Pop$, there are at least two individuals $m,n\in \Pop$ with phenotypical distance at least $\eps|\Cases|$.

If instead $\Cases$ is a training set with real-valued losses, then let $\eps \in [0,1]$ and $\delta >0$. Then the \emph{$\eps$-Cluster Similarity for $\delta$-distance} is defined as the minimal $k\ge 2$ such that among every set of $k$ different individuals in $\Pop$, there are at least two individuals $m,n\in \Pop$  with phenotypical $\delta$-distance at least $\eps|\Cases|$.
\end{definition}

A few remarks are in order to understand the definition better. 
Firstly, the $\eps$-Cluster Similarity $k$ is a \emph{decreasing} measure of diversity, i.e., less similarity means more diversity and vice versa. 
Moreover, the value $k$ is increasing in $\eps$: we are only satisfied with two individuals of distance at least $\eps|\Cases|$, which is harder to achieve for larger values of $\eps$. 
Therefore, we may need a larger set to ensure that it contains a pair of individuals with such a large distance. 
In other words, a larger value of $\eps$ means that we are more restrictive in counting individuals as different, which yields larger value of $k$: the population is more similar with respect to a more restrictive measure of difference. 
There is an important tradeoff between $\eps$ and $k$: larger values of $\eps$ (which are desirable in terms of diversity, since we search for individuals with larger distances) lead to larger values of $k$ (which is undesirable since we only find such individuals in larger sets). 

Second, having small $\eps$-Cluster Similarity is a rather weak notion of diversity: it does not require that \emph{all} pairs of individuals are different from each other. For example, if the population consists of clusters of $k-1$ individuals which are pairwise very similar, 
then the $\eps$-Cluster Similarity is $k$ as long as the clusters have distances at least $\eps|\Cases|$ from each other. We just forbid that there is a cluster of size $k$ such that \emph{every} pair of individuals in the cluster has small distance.

On the other hand, the $\eps$-Cluster Similarity may be a finer measure than, say, the average phenotypical distance in the population. For example, consider a population that consists only of two clusters of almost identical individuals, but the clusters are in opposite corners of the phenotype space, i.e., they differ on almost all training cases. Then the average phenotypical distance is extremely large, $\approx |\Cases|/2$, which would suggest high diversity. But even for absurdly high $\eps = 0.9$, we would find $k= |\Pop|/2 + 1$, i.e., a very low diversity according to our definition. It is not hard to see that in this example the expected running time of lexicase selection is $\Omega(|\Pop|\cdot|\Cases|)$: in the first step one of the clusters will be removed completely, but afterwards it is very hard to make any further progress. Hence, this example shows that \textit{average phenotypical distance does not predict the running time of lexicase selection well}: even though the example has large average phenotypical distance (``large diversity'' in that sense), the running time of lexicase selection is very high. The main theoretical insight of this paper is that this discrepancy can never happen with $\eps$-Cluster Similarity. Whenever $\eps$-Cluster Similarity is low (large diversity), then the expected running time of lexicase selection is small.\medskip

To give the reader another angle to grasp the definition of $\eps$-Cluster Similarity, we give a second, equivalent definition in terms of graph theory.

\begin{definition}[$\eps$-Cluster Similarity, Equivalent Definition]\label{def:diversity2}
Let $\Pop$ be a population of search points, and $\Cases$ be a set of training cases with discrete losses. Let $\eps \in [0,1]$. We define a graph $G=(V,E)$ as follows. The vertex set $V := \Pop$ is identical with the population. Between any two vertices $m,n\in \Pop$, we draw an edge if and only if the individuals $m$ and $n$ have the same loss in more than $(1-\eps)|\Cases|$ training cases. Then the \emph{$\eps$-Cluster Similarity} is $k := \alpha +1$, where $\alpha$ is the clique number of $G$, i.e, $\alpha$ is the size of the largest clique of the graph $G$.

If $\Cases$ is a training set with real-valued losses and $\delta >0$ a parameter, then we use the same vertex set for $G = G(\delta)$, but we draw an edge between $m$ and $n$ if and only if the losses of $m$ and $n$ differ by at most $\delta$ for more than $(1-\eps)|\Cases|$ training cases. Then the \emph{$\eps$-Cluster Similarity for $\delta$-distance} is again $k := \alpha +1$, where $\alpha$ is the clique number of $G(\delta)$.
\end{definition}

\section{Theoretical Result: Low $\eps$-Cluster Similarity leads to small running times}\label{sec:running-time}

In this section, we prove mathematically, that a high diversity (i.e., a small $\eps$-Similarity) leads to a small expected running time for lexicase selection.

\subsection{Preliminaries}

To proof our main theoretical result, we will use the following theorem, known as Multiplicative Drift Theorem~\cite{doerr2012multiplicative,lengler2020drift}, which is a standard tool in the theory of evolutionary computation.

\begin{theorem}[Multiplicative Drift]
\label{driftthm:multiplicative}
Let $(X_t)_{t\geq 0}$ be a sequence of non-negative random variables with a finite state space $\mathcal{S} \subseteq \R_0^+$ such that $0 \in \mathcal S$. 
Let $s_{\min} := \min(\mathcal{S} \setminus \{0\})$, let $T := \inf\{t \geq 0 \mid X_t =0\}$, and for $t \geq 0$ and $s \in \mathcal S$ let $\Delta_t(s) := E[X_{t} - X_{t+1} \mid X_t = s]$. 
Suppose there exists $\delta >0$ such that for all $s\in \mathcal{S}\setminus\{0\}$ and all $t\geq 0$ the drift is
\begin{align}\label{drifteq:multdrift1}
\Delta_t(s) \geq \delta s.
\end{align}
Then
\begin{align}\label{drifteq:multdrift2}
E[T] \leq \frac{1+E[\ln(X_0/s_{\min})]}{\delta}.
\end{align}
\end{theorem}

Now we can give our theoretical results. Note that the following theorems refer to a single execution of lexicase selection, i.e., $M$ refers to the complexity of finding a single parent via lexicase selection. The following theorem says that the running time is low, $O(|\Pop|+|\Cases|)$ if the population has large $\eps$-Cluster Similarity. As common in theoretical running time analysis, we give the running time in terms of \emph{evaluations}, where an evaluation is an execution (or lookup) of $L(n,c)$ for an individual $n$ and a training case $c$. The running time is proportional to the number of evaluations.

\begin{theorem}\label{thm:clusters}
Let $0<\eps<1$. Consider lexicase selection on a population $\Pop$ without duplicates and with $\eps$-Cluster Similarity of $k \in \N$. Let $M$ be the number of evaluations until the population pool is reduced to size $1$. 
Then
\[
E[M] \le \frac{4|\Pop|}{\eps} + 2k |\Cases|.
\]
\end{theorem}
\begin{proof}
Consider any two individuals $m,n\in \Pop$. Assume that both $m,n$ are still in the candidate pool after some selection steps (i.e., after some iterations of the while-loop have been processed). Then $m$ and $n$ can not differ in any of the processed cases $\Cases\setminus\Cases'$, because otherwise one of them would have been removed from the population. Therefore, the $\eps|\Cases|$ cases in which $m$ and $n$ differ are all still contained in $\Cases'$. In particular, if we choose a new case from $\Cases'$ at random, then the probability that $m$ and $n$ differ in this case is at least $\eps|\Cases|/|\Cases'| \geq \eps$. Note that this holds throughout the algorithm and for any two individuals $m,n$ that are still candidates.

%
Now we turn to the computation. Let $X_t$ be the number of remaining individuals after $t$ executions of the while-loop. We define $Y_t := X_t$ if $X_t \geq 2k$ and $Y_t := 0$ if $X_t < 2k$. Let $T'$ be the first point in time when $Y_{T'}=0$ (and thus, $X_{T'} < 2k$). 

If $X_t \geq 2k$, then we split the population before the $t+1$-st step into pairs as follows. Since $X_t \ge k$, there are at least two individuals which differ in at least $\eps|\Cases|$ cases, so we pick two such individuals and pair them up. We can iterate this until there are less than $k$ unpaired individuals left. Therefore, we are able to pair up at least $X_t- (k-1) > X_t - X_t/2 = X_t/2$ individuals, forming at least $X_t/4$ pairs. For each pair, there is a chance of $\eps$ that the two differ in the case of the $t+1$-st step, in which case at least one of them is eliminated. Hence, for every $x\geq 2k$,
\[
E[X_{t+1} \mid X_t = x] \leq x -\tfrac{\eps x}{4} = x(1-\tfrac\eps4).
\]
Now let us assume that $Y_t = y >0$ (and thus $y\ge 2k$). Since $Y_{t+1} \le X_{t+1}$ by definition, we obtain
\begin{align}\label{eq:pairwise-distance-2}
E[Y_{t+1} \mid Y_t = y] & \le E[X_{t+1} \mid X_t = y] \leq   y(1-\tfrac\eps4).
\end{align}
The advantage of $Y_t$ is that the above bound holds for all $y\geq 0$ (it is trivial for $y=0$), whereas the corresponding bound for $X_t$ may not hold for $0 < x< 2k$. 

Now we bound $M$ by splitting it into the running time $M_1$ before step $T'$, and the running time $M_2$ after and including step $T'$. For $M_1$, we proceed as follows. 
\[
M_1 = \sum\nolimits_{t=0}^{T'-1} X_t = \sum\nolimits_{t=0}^{T'-1} Y_t = \sum\nolimits_{t=0}^{\infty} Y_t,
\]
because $Y_t = 0$ for $t\geq T'$. Applying~\eqref{eq:pairwise-distance-2} iteratively to $Y_t$, we obtain 
\[
E[Y_t] \leq (1-\tfrac\eps4)^t Y_0,
\]
where $Y_0 = |\Pop|$. Plugging this in, we get
\begin{align*}
E[M_1] & = \sum\nolimits_{t=0}^{\infty} E[Y_t] \le  \sum\nolimits_{t=0}^{\infty}(1-\tfrac\eps4)^t(|\Pop|-1) \le \tfrac{4|\Pop|}{\eps},
\end{align*}
where in the last step we have used the formula $\sum_{t=0}^\infty q^t = 1/(1-q)$ for geometric series with $q=1-\eps/4$. It remains to bound $M_2$, and we use a simple bound. Since every case occurs at most once, and since the population size is at most $2k$, we have deterministically $M_2 \le2 k\cdot |\Cases|$.
\end{proof}

\section{Empirical Evaluation in Program Synthesis}

We evaluated the theoretical bounds given by \cref{thm:clusters} on examples using genetic programming to solve program synthesis benchmark problems. 
The purpose of this evaluation is to 1) find out how diverse the populations are according to $\eps$-Cluster Similarity; 2) measure the extent to which the new bounds shrink our estimates of the running time of lexicase selection, relative to the known worst-case bounds; 3) evaluate the sensitivity of $\eps$-Cluster Similarity to the parameter, $\eps$, across several problems; and 4) determine how $\eps$-Cluster Similarity compares to a more standard diversity metric in real data.

\subsection{Experimental Setup}

We investigate these aims using 8 program synthesis problems taken from the General Program Synthesis Benchmark Suite~\cite{Helmuth:2015:GECCO}. 
These problems require solution programs to manipulate multiple data types and exhibit different control structures, similar to the types of programs we expect humans to write. Among the 8 problems there are 5 different expected output types (Boolean, integer, float, vector of integers, and string), allowing us to test against multiple data types. In particular, we note that two problems (compare-string-lengths and mirror-image) have Boolean outputs, which we expect to have higher $\eps$-Cluster Similarity values due to having fewer possible output values.

Our experiments were conducted using PushGP, which evolves programs in the Push programming language~\cite{spector:2002:GPEM,1068292}. PushGP is expressive in the data types and control structures it allows, and has been used previously with these problems~\cite{Helmuth:2015:GECCO}. We use the Clojush, the Clojure implementation of Push, in our experiments.\footnote{\url{https://github.com/lspector/Clojush}}
Each run evolves a population of 1000 individuals for a maximum of 300 generations using lexicase selection and UMAD mutation (without crossover)~\cite{Helmuth:2018:GECCO}. We conduct 100 runs of genetic programming per problem.

For each problem, trial, generation, and selection event, we calculated the $\eps$-Cluster Similarity for $\eps \in [0.05, 0.6]$ in increments of 0.05. 
Using these values, we calculated the bound on the expected running time of lexicase selection according to \cref{thm:clusters}.
For comparison, we also calculated 1) the worst-case complexity of lexicase selection at those operating points and 2) the average pair-wise covariance of the population error vectors.
We calculated worst-case running time as $N \cdot C$, neglecting constants, in order to make our comparison to the new running time calculation conservative.

\subsection{Results}

\begin{figure}[!ht]
    \includegraphics[width=\textwidth]{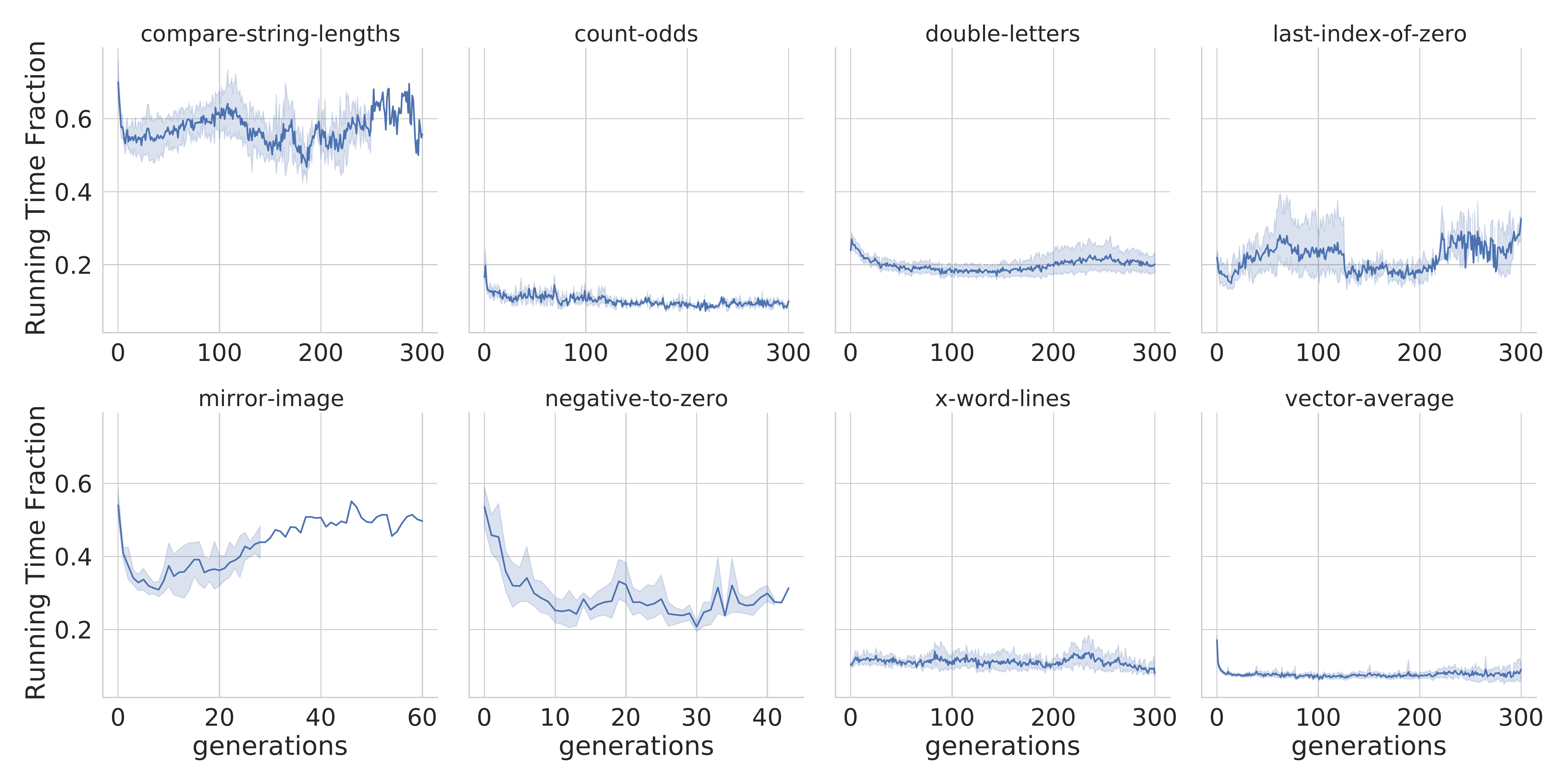}
    \caption{
        New running time bound divided by the previously known worst-case bound, as a function of evolutionary generations. 
        The $y$-axis shows the ratio of both bounds using measurements of relevant parameters.
        The filled region represents confidence interval of the estimates over all trials.  
    }
    \label{fig:running-time-frac}
\end{figure}

\cref{fig:running-time-frac} visualizes the new running time bound as a fraction of the bound given by the worst-case complexity, $N\cdot C$. 
Across problems, the running time bound given by \cref{thm:clusters} ranges from approximately 10-70\% of the worst-case complexity bound, indicating much lower expected running times. 
On average over all problems, the bound given by \cref{thm:clusters} is 24.7\% of the worst-case bound on running time.

\begin{figure}[!ht]
    \includegraphics[width=\textwidth]{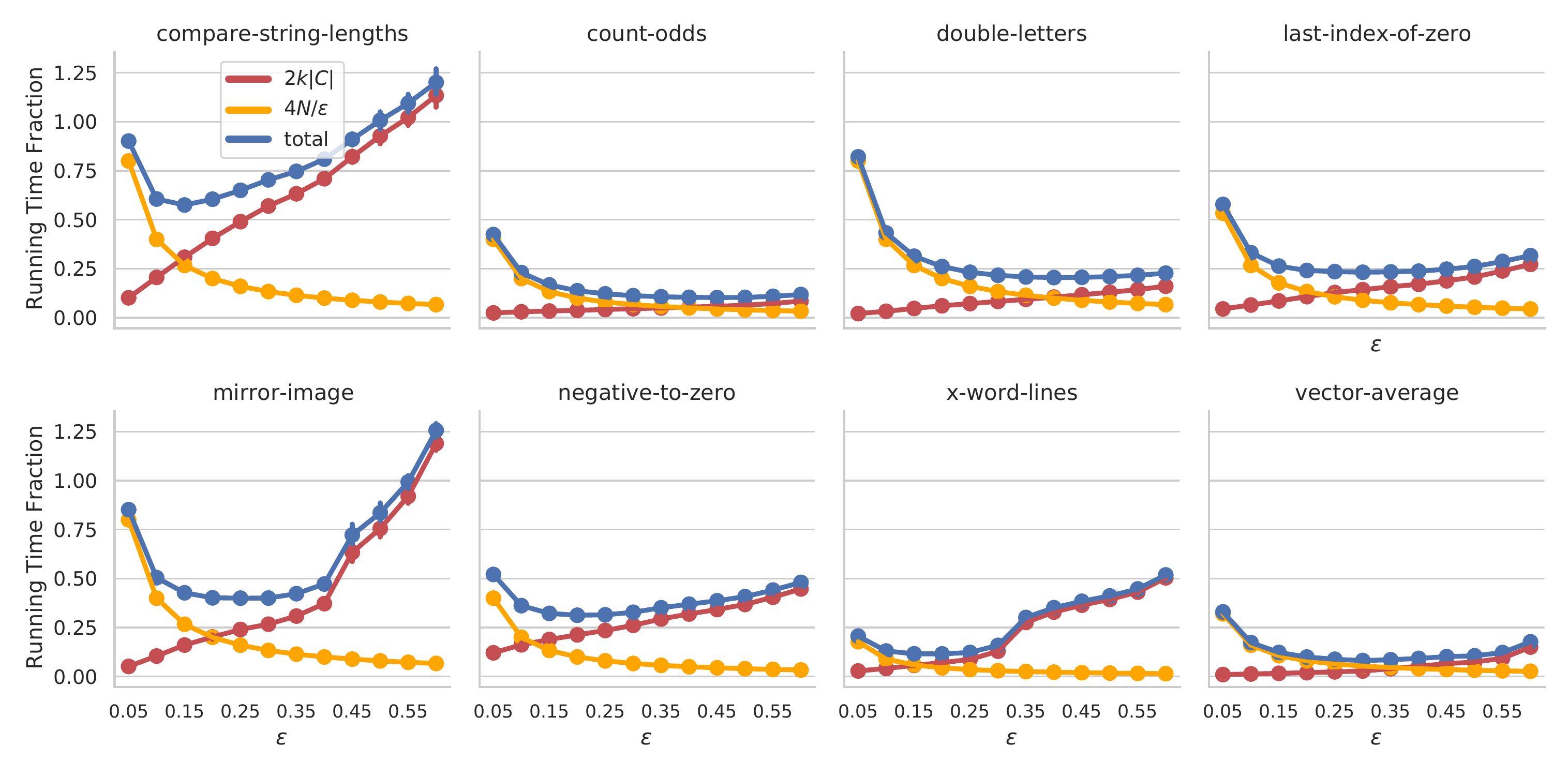}
    \caption{
        The ratio of new running time bound and previous known worst-case bound as a function of $\eps$. 
        Optimal values vary by problem but we note flat regions for many problems suggesting a broad range of possible $\eps$ values that give similar running time bounds. 
    }
    \label{fig:best-eps}
\end{figure}

\cref{fig:best-eps} shows the components of \cref{thm:clusters} as a function of $\eps$, as well as the total expected running time bound.
For small values of $\eps$, the $4|\Pop|/\eps$ term dominates, whereas for larger values of $\eps$, the $2k|\Cases|$ term dominates. 
The observed behavior agrees with our intuition, since larger values of $\eps$ lead to larger values of $k$.
The value of $\eps$ corresponding to the lowest bound on running time varies by problem, with an average value of 0.29. 

\begin{figure}[!ht]
    \includegraphics[width=\textwidth]{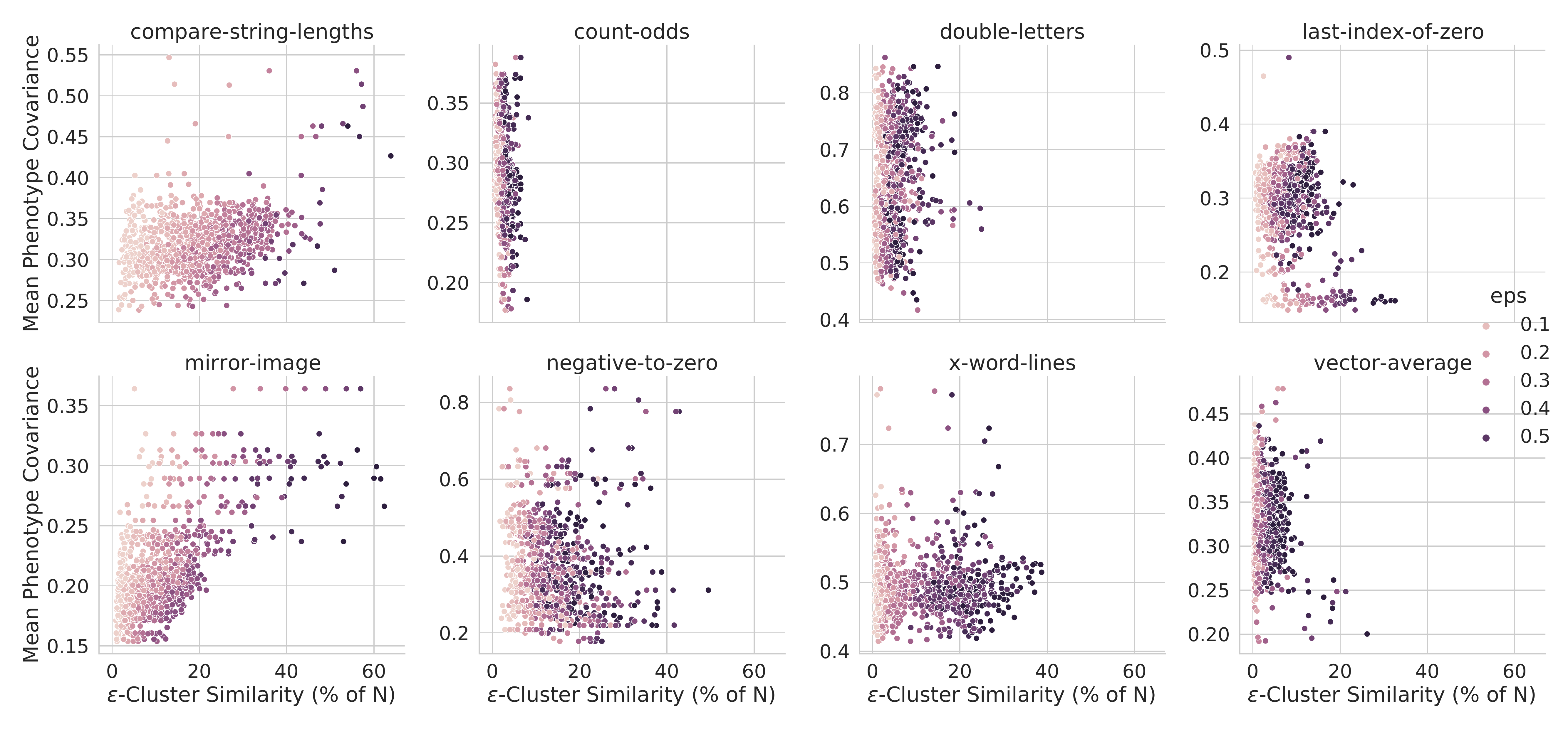}
    \caption{
        A comparison of $\eps$-Cluster Similarity (x-axis) and the mean of the covariance matrix of population error (y-axis), colored by $\eps$.  
        $\eps$-Cluster Similarity ($k$) is plotted as a percent of the population size, $N$. 
        In most cases (6/8), we observe little relation between the two measures, suggesting $\eps$-Cluster Similarity is indeed measuring a distinct aspect of population diversity, one that is particularly relevant to the running time of lexicase selection.
    }
    \label{fig:cov}
\end{figure}

\cref{fig:cov} compares the new diversity metric (\cref{def:diversity2}) to a more typical definition of behavioral diversity: the mean of the covariance matrix given by population errors. 
In general, we observe that $\eps$-Cluster Similarity does not correlate strongly with mean covariance, suggesting that it does indeed measure a different aspect of phenotypic diversity as suggested in \cref{sec:eps_cluster}.

\section{Discussion}

We see in Fig.~\ref{fig:running-time-frac} that the new running time bound is below the old bound, and sometimes substantially lower by a factor of $5-10$. Thus a substantial part of the discrepancy between the (old) worst-case running time bound and the empirically observed fast running times can be explained by the fact that populations in real-world data are diverse according to our new measure. Since the theoretical analysis is still a worst-case analysis (over all populations), we do not expect that the new bound can explain the whole gap in all situations, but it does explain a substantial factor.

In Fig.~\ref{fig:best-eps}, we investigate which choice of $\eps$ gives the best running time bound. Note that $\eps$ is a parameter that can be chosen freely. Every choice of $\eps$ gives a value $k$ for the $\eps$-Cluster Similarity, which in turn gives a running time bound. While in Fig.~\ref{fig:running-time-frac} we plotted the best bound that can be achieved with any $\eps$, Fig.~\ref{fig:best-eps} shows for each $\eps$ the bound that can be obtained with this $\eps$. It is theoretically clear that the term $4N/\eps$ (yellow) is decreasing in $\eps$ and $2kC$ (red) is increasing (since $k$ increases with $\eps$). The bound (blue) is the sum of these two terms, and we observe that very small and very large choices of $\eps$ often give less good bounds. However, often the blue curve shows some range in which it is rather flat, indicating that there is a large range of $\eps$ that gives comparable bounds. In particular, it seems that the range $0.15 \le \eps \le 0.25$ often gives reasonable bounds.

In Fig.~\ref{fig:cov} we compare the $\eps$-Cluster Similarity ($k$ normalized as a percent of $N$) with another diversity measure: the mean of the covariance matrix of population error. If both measures of diversity were highly correlated, we would expect that for any fixed $\eps$ (points of the same color), points with larger $x$-value would consistently have smaller $y$-value (since $k$ is an inverse measure of diversity). However, in many plots this correlation is spurious at best. It even appears to have opposite signs in some cases. We conclude that $\eps$-Cluster Similarity measures a different aspect of diversity than the mean of the covariance matrix. 
We also note that, interestingly, the two problems for which there \textit{does} appear to be a relation between the two measures (compare-string-lengths and mirror-image) are the two problems with boolean error vectors. 

\section{Conclusions}

We have introduced and investigated a new measure of population diversity, the $\eps$-Cluster Similarity. We have theoretically proven that large population diversity makes lexicase selection fast, and empirically confirmed that populations are diverse in real-world examples of genetic programming. Thus we have concluded that diverse populations can explain a substantial part of the discrepancy between the general worst-case bound for lexicase selection, and the fast running time in practice. 

Naturally, the question arises whether populations in other areas than genetic programming are also diverse with respect to this measure. Moreover, what other consequences does a diverse population have? For example, does it lead to good crossover results? Does it help against getting trapped in local optima? While the intuitive answer to these questions seems to be Yes, it is not easy to pinpoint such statements with rigorous experimental or theoretical results. We hope that our new notion of population diversity can be a means to better understand such questions. 

\subsubsection{Acknowledgements} 
William La~Cava was supported by the National Library of Medicine and National Institutes of Health under award R00LM012926. We would like to thank Darren Strash for discussions that contributed to the development of this work.

%
%

\bibliographystyle{splncs04}
\bibliography{lexreferences,bills_refs}





\end{document}